%% file: main.tex
\documentclass{article}

\PassOptionsToPackage{numbers, compress}{natbib}
 \usepackage[preprint]{neurips_2026}

\usepackage[utf8]{inputenc} 
\usepackage[T1]{fontenc}    
\usepackage{hyperref}       
\usepackage{url}            
\usepackage{booktabs}       
\usepackage{amsfonts}       
\usepackage{nicefrac}       
\usepackage{microtype}      
\usepackage[table]{xcolor}         
\usepackage{graphicx}
\usepackage{amsmath,amssymb}
\usepackage{multirow,multicol}
\usepackage{array}
\usepackage{wrapfig}

\title{On the Implicit Reward Overfitting and the Low-rank Dynamics in RLVR}

\definecolor{headercolor}{RGB}{236, 247, 236} 
\definecolor{basiccolor}{RGB}{242, 243, 248}   
\definecolor{complexcolor}{RGB}{255, 245, 220} 

\author{%
Hao Ye$^{1}$~~
Jisheng Dang$^{1}$\thanks{Corresponding Author}~~
Junfeng Fang$^{2*}$~~
Bimei Wang$^{1}$~~ 
Yizhou Zhang$^{1}$~~
Ning Lv$^{1}$ \\
Wencan Zhang$^{2}$~~
Hong Peng$^{1}$~~
Bin Hu$^{3*}$~~
Tat-Seng Chua$^{2}$~~ \\
$^{1}$ Lanzhou University \
$^{2}$ National University of Singapore \
$^{3}$ Beijing Institute of Technology
}

\begin{document}

\maketitle

\begin{abstract}
Recent extensive research has demonstrated that the enhanced reasoning capabilities acquired by models through Reinforcement Learning with Verifiable Rewards (RLVR) are primarily concentrated within the rank-1 components. Predicated on this observation, we employed Periodic Rank-1 Substitution and identified a counterintuitive phenomenon: RLVR may exhibit implicit reward overfitting to the training dataset. Specifically, the model can achieve satisfactory performance on the test set even when its rewards remain relatively low during the training process. Furthermore, we characterize three distinct properties of RL training: (1) The effective rank-1 component in RLVR don't maintain other model knowledge except mathematical reasoning capability. (2) RLVR fundamentally functions by optimizing a specific singular spectrum. The distribution of singular values of almost all linear layers in RLVR-trained model behaves like heavy-tailed distribution. (3) the left singular vectors associated with rank-1 components demonstrate a stronger alignment tendency during training, which echoes the discovery that RLVR is optimizing sampling efficiency in essence. Taken together, our findings and analysis further reveal how RLVR shapes model parameters and offer potential insights for improving existing RL paradigms or other training paradigms to implement continual learning. 
\end{abstract}

\input{1-introduction}
\input{2-methods}
\input{3-experiment}
\input{4-related-work}
\input{5-conclusion}

\clearpage
\bibliographystyle{plainnat}
\bibliography{reference}





\newpage
\input{6-appendix}



\end{document}

%% file: 1-introduction.tex
\section{Introduction}

In recent years, Reinforcement Learning (RL) \cite{schulman2015trpo,schulman2017proximal,rafailov2023direct,ahmadian2024back} has emerged as a core paradigm for aligning Large Language Models (LLMs)~\cite{team2024qwen2} with human preferences and, more importantly, for eliciting complex reasoning capabilities within these models. From early methods like RLHF~\cite{ouyang2022-instructgpt-rlhf} to recent reasoning-oriented models such as DeepSeek-R1~\cite{guo2025deepseek-r1} and Kimi-K2~\cite{kimiteam2025kimik2openagentic}, RL has demonstrated remarkable potential. However, despite the community's immense success in engineering practices, our understanding of how RL actually alters model parameters remains underdeveloped. Existing research~\cite{sequeira2020interestingness,heuillet2022collective} predominantly focuses on macroscopic training dynamics, interaction to agents or reward design, while few studies delve into the microscopic level to investigate the specific structure and physical significance of RL updates $\Delta W$ within the parameter space.

This lack of insight into microscopic mechanisms directly contributes to our confusion regarding the boundaries of RL capabilities. Recently, a substantial body of researches~\cite{zhang2024lolcats} have observed that LLMs exhibit low-rank characteristics when adapting to downstream tasks, and RL appears to be no exception. A recent study~\cite{cai2025predictability} suggests that the reasoning improvements conferred by RL are primarily concentrated within the Rank-1 component of weight updates. Closely related work has boldly proposed the universal weight subspace hypothesis~\cite{kaushik2025universal}. Furthermore, another disruptive study~\cite{yue2025limit-of-rlvr} recently found that even without RL training, the pass@k~\cite{chen2025passatk} metric of models continues to improve as $k$ increases. This implies that RL may not be imparting new underlying reasoning logic to the model, but rather optimizing its sampling strategy to efficiently elicit latent correct answers. This perspective has triggered profound questioning within the community regarding the true nature of RLVR: is RLVR genuinely learning reasoning, or is it merely fitting the reward?
\begin{figure*}[t]
    \centering
    \includegraphics[width=\linewidth]{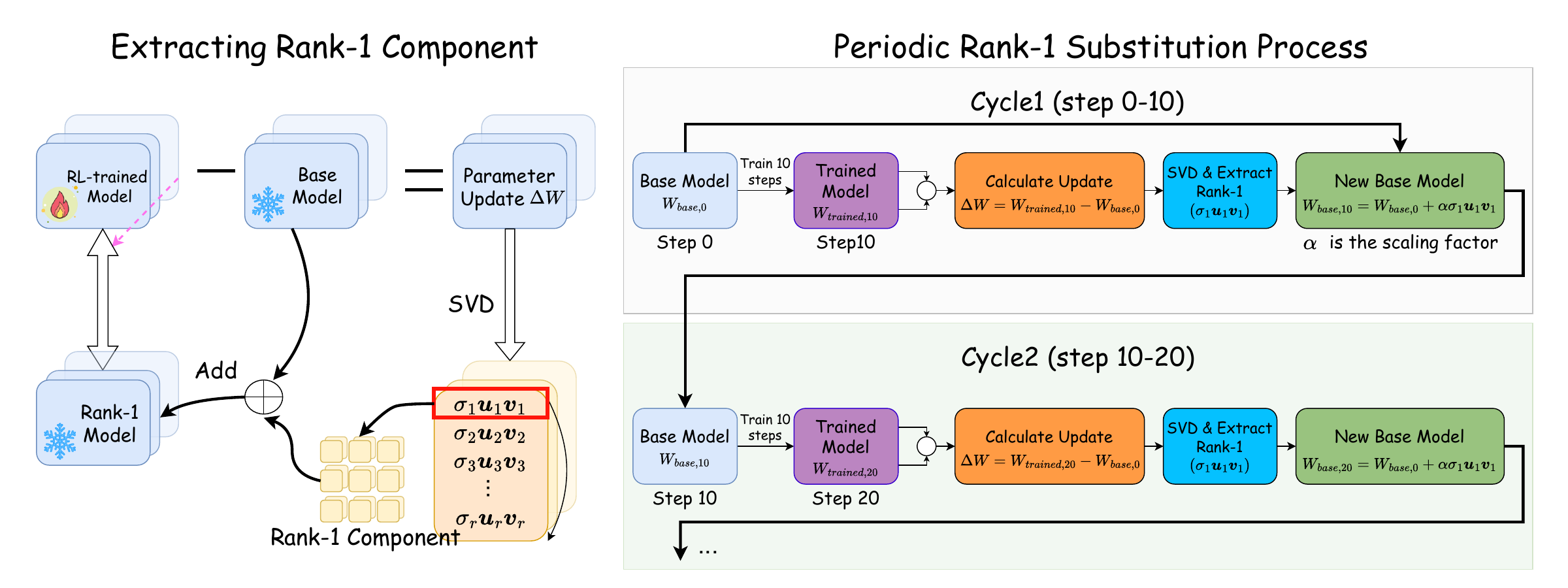}
    \caption{\textbf{Left}: The process of extracting Rank-1 component from RL-trained model. We pick out the rank-1 matrix corresponding to the greatest singular value. \textbf{Right}: The process of periodic Rank-1 substitution. base model is repeatedly trained for a short interval, and only a rank-1 approximation of its weight update is kept.}
    \label{fig:fig1}
\end{figure*}

Another point of interest is the low-rank property of RLVR. A research from Thinking Machines~\cite{schulman2025lorawithoutregret} found that LoRA performs equivalently to full fine-tuning for reinforcement learning even with small ranks. They find that RL requires very low capacity, a result based on information-theoretical arguments. This observation echoes the fact mentioned above that the rank-1 component of parameter update obtained in RLVR contains most of the reasoning capability~\cite{cai2025predictability}. Therefore, it's significant to further explore how LoRA learns this rank-1 component under the hood.

To unravel these mysteries, this paper conducts an in-depth empirical study and theoretical analysis starting from the spectral properties of weight updates. Our work reveals three critical phenomena during the RLVR training process:

First, we identify that RLVR may suffer from Implicit Reward Overfitting. By introducing a technique called Periodic Rank-1 Substitution, we discover that a vast number of non-Rank-1 components in RLVR (e.g., GRPO~\cite{shao2024deepseekmath}, DAPO~\cite{yu2025dapo}, GSPO~\cite{zheng2025gspo}) do not contribute to the improvement of reasoning capabilities. By periodically eliminating these components, we find that on a significant portion of datasets, the model's performance on the test set remains comparable, even though its training reward is lower than that of Full RL Fine-tuning. This indicates that while non-Rank-1 components significantly boost the reward on the training set, they fail to translate into generalized performance on the test set. Furthermore, due to the nature of the reward function in the RL process, the model may be overfitting to the training data.

However, if the rank-1 component is enough to obtain equivalent reasoning improvement, what's the side-effect of discarding non-rank-1 component during training? Thus, we further observed the effect of non-rank-1 components and corroborated that non-rank-1 components maintain the model's out-of-domain capability, such as world knowledge, instruction-following ability and safety, etc. Besides, we analyzed the singular spectra (the distribution of singular values) of all linear layers in RLVR-tuned models and found that they follow a RL-induced pattern: a leading spike followed by a heavy-tailed distribution.

Third, we explain the geometric dynamics of LoRA in RL: the Geometric Asymmetry of Input-Output Subspaces. We find that weight updates of almost all linear layers induced by RLVR exhibit strong Rank-1 characteristics on the output side (Output Space), which easily align with the left singular vectors (corresponding to adjustments in sampling strategy) of the greatest singular value. However, on the input side (Input Space) , updates involve complex reorganization of contextual features, which are difficult to capture with simple low-rank matrices. This observation also reflects the fact that RLVR is inherently optimizing sampling(output) efficiency.

To summarize, the contributions of our work are:
\begin{itemize}
    \item We reveal \textbf{potential implicit overfitting} in RLVR, and we give a plausible explanation to a disruptive discovery that reinforcement learning don't incentivize reasoning capacity in LLMs beyond the base model~\cite{yue2025limit-of-rlvr}.
    \item We find that non-rank-1 components in RLVR training maintain the overall knowledge and ability of model except reasoning ability.
    \item We discover that the singular spectrum of the parameter update of a RL-tuned model follow a consistent pattern: a leading spike followed by a heavy-tailed distribution.
    \item We provide an in-depth analysis of the training dynamics of LoRA in RLVR: the asymmetry of the input-output space.

\end{itemize}

%% file: 2-methods.tex
\section{Rank-1 Dominance in RL: Decoupling underlying Reasoning from Implicit Reward Overfitting}
\label{rank1-preliminary}
\subsection{Preliminary: The Dominance of Rank-1 Subspace}

Suppose $\Delta W$ is the parameter updates after RL training. We perform SVD on $\Delta W$:
\begin{gather*}
    \Delta W=\sum_{i=1}^r \sigma_i\boldsymbol{u}_i\boldsymbol{v}_i^T
\end{gather*}
where $\sigma_i$ are singular values in descending order and $\boldsymbol{u}_i$,$\boldsymbol{v}_i$ are left and right singular vectors. The \textbf{Rank-1 Component} (or \textbf{Rank-1 Subspace}) denotes the rank-1 matrix obtained from the greatest singular value:
\begin{gather*}
    \Delta W^{(1)}=\sigma_1\boldsymbol{u}_1\boldsymbol{v}_1^T
\end{gather*}
To ensure consistency in update strength, this Rank-1 component should be rescaled by Frobenius norm to match the magnitude of the original update:
\begin{gather*}
    \Delta \hat{W}^{(1)}=\dfrac{\Vert\Delta W\Vert_F}{\Vert\Delta W^{(1)}\Vert}_F \Delta W^{(1)}
\end{gather*}
The Rank-1 model is obtained by adding $\Delta \hat{W}^{(1)}$ to base model. Existing study found that the rank-1 model boasts virtually equivalent reasoning capability compared to model fully fine-tuned by RL~\cite{cai2025predictability} and ablation study demonstrated that cumulative rank-k components performs almost equally to the rank-1 component as k grows.

\subsection{Periodic Rank-1 Substitution: Eliminating Non-rank-1 Component}
Extensive empirical evidence from prior studies suggests that the enhancement in reasoning capabilities elicited by RLVR is predominantly encapsulated within the Rank-1 component of the weight updates. This observation naturally induces a critical question: within the RL weight update matrix $\Delta W$, do the residual non-Rank-1 components merely constitute stochastic noise inherent to the training process? 

To empirically interrogate this noise hypothesis, we conducted comparative training utilizing both standard GRPO and a variant we term GRPO with \textbf{Periodic Rank-1 Substitution} on the Countdown-3to4 dataset. As illustrated in Figure \ref{fig:fig1}, our proposed Periodic Rank-1 Substitution mechanism operates as follows: at intervals of 10 training steps, we perform Singular Value Decomposition (SVD) on the accumulated weight updates relative to the base model. Subsequently, the extracted Rank-1 component is added onto the base parameters, serving as the initialization for the subsequent 10-step training window. The remaining training steps proceeds likewise.
\begin{figure*}[ht]
    \centering
    \includegraphics[width=\linewidth]{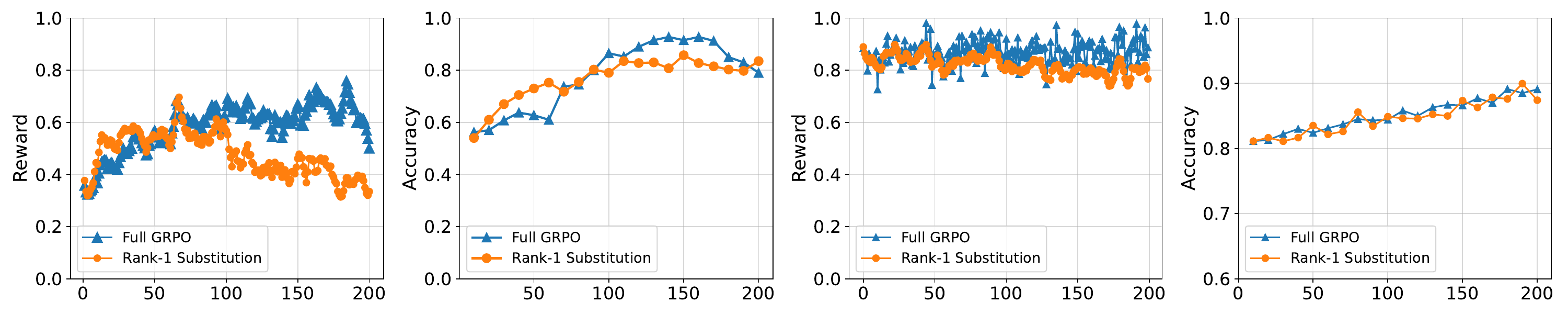}
    \caption{\textbf{Left}: The mean reward within each batch during GRPO training for Qwen2.5-7B-Instruct~\cite{qwen2025qwen25technicalreport}. \textbf{Mid-left}: Test-set accuracy of the leftmost figure. \textbf{Mid-right}: The mean reward within each batch during GRPO training for Llama3.1-8B-Instruct~\cite{grattafiori2024llama3herdmodels}. \textbf{Right}: Test-set accuracy of the Mid-right figure. The horizontal axes of four subfigures are training steps. We don't use Qwen3 as post-training has enabled the lightweight models in the Qwen3 family to acquire reasoning capabilities distilled from larger, RL-trained models, as described by the Qwen3 tech reports~\cite{yang2025qwen3technicalreport}, which makes it harder to measure what is being learned only during RLVR.}
    \label{fig:fig2-overfit}
\end{figure*}

\subsection{Divergence between Training Reward and Test Generalization}
Figure \ref{fig:fig2-overfit} presents the trajectories of the training mean reward alongside the evaluation metrics on the test set across varying training steps for both strategies. We observe a counterintuitive and prominent mismatch between the training reward and testing performance: while standard GRPO exhibits a consistent upward trend and achieves significantly higher training rewards compared to the Periodic Rank-1 Substitution variant, this advantage does not translate into a superiority in test performance, where only a marginal gap is observed.

A recent study posits that RLVR does not genuinely elicit the model's underlying reasoning ability. Rather, it primarily optimizes the sampling strategy to efficiently prioritize correct solutions. Consequently, due to the reduced diversity in sampling post-RL, the model may fail to answer a subset of queries that the base model could originally solve, particularly when evaluated under larger $k$ values for pass@k. 

Our empirical findings provide robust support for this assertion: satisfactory evaluation metrics on the test set do not preclude the possibility of overfitting to the training data. Specifically, existing RLVR paradigms are potentially prone to implicit reward overfitting. To put it in another perspective, there exists a potential risk of over-optimizing the sampling strategy during the RL process. This elucidates why, as Pass@k increases, the performance of the standard RL model may unexpectedly fall short of the base model~\cite{dang2025assessing}. By extracting the weight components responsible for reasoning and filtering out non-Rank-1 noise, our Periodic Rank-1 Substitution strategy effectively mitigates this implicit reward overfitting while retaining the desired target weight updates for reasoning performance, thereby preserving generalization capabilities. From the perspective of learning dynamics, eliminating the noise that artificially inflates training rewards forces the model to concentrate on core reasoning mechanisms. Even with a weaker reward signal, the model successfully acquires the critical logic for problem-solving through calibration along these principal directions.
\begin{wrapfigure}{r}{0.5\textwidth}
    \centering
    \vspace{-2mm}
    \!\!\!\!\!\!\!\!\includegraphics[width=0.49\textwidth]{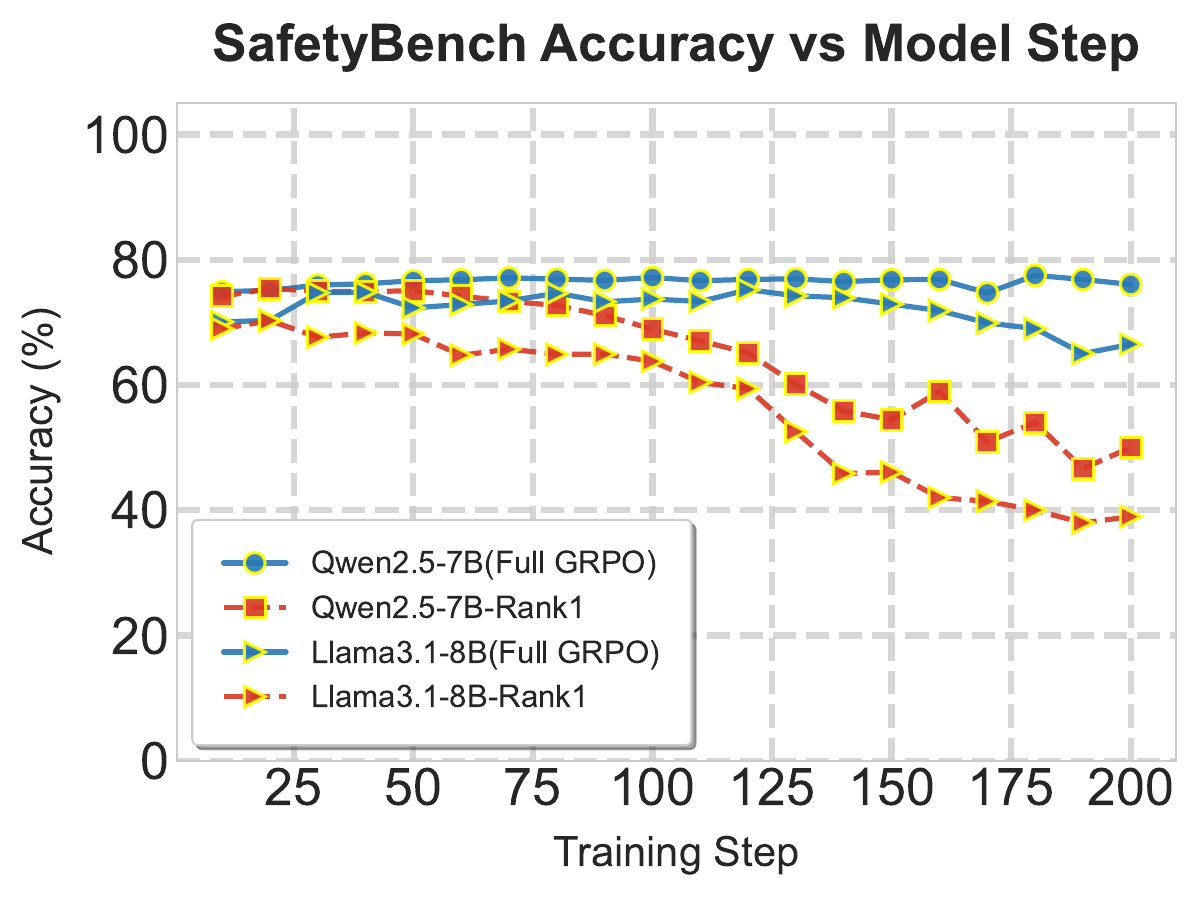}
    \caption{Obvious performance degradation in safety after RLVR with periodic rank-1 subsitution. Models are the same as those in Figure~\ref{fig:fig2-overfit}}
    \vspace{-4mm}
    \label{fig:fig4-safety}
    \vspace{-8mm}
\end{wrapfigure}
\section{Non-rank-1 Components are Better Than Nothing}

\subsection{Non-rank-1 Components are NOT Functionally Redundant}

As illustrated in Figure \ref{fig:fig4-safety}, we evaluated the performance of both the fully RLVR-tuned model and the model fine-tuned by RLVR with periodic rank-1 substitution on the SafetyBench benchmark. We observed a prominent degradation in safety metrics following the Rank-1 extraction. This suggests that while the non-Rank-1 components contribute minimally to the enhancement of reasoning capabilities, they may encode other behavioral shifts acquired during the RL process. 

To further investigate the role of non-rank-1 components during the RLVR process, and taking into account that the results shown in Figure~\ref{fig:fig4-safety} might be attributed to frequent manual parameter updates applied to the model during training, we conducted an additional comparative experiment. To ensure generalizability, we analyzed the performance of LLMs with varying parameter sizes, from different model families, trained on diverse datasets using various reinforcement learning methods. Specifically, we evaluated their performance on non-mathematical reasoning benchmarks before and after rank-1 extraction. Here, the rank-1 extraction is only performed once right after the whole training process. For example, Llama3.1-8B-Thinking-R1-rank1 is obtained by adding rank-1 component to Llama3.1-8B-Instruct~\cite{grattafiori2024llama3herdmodels}. As shwon in Table~\ref{tab:tab1-effect-of-non-rank1}, there are two key observations:

\paragraph{Non-rank-1 components maintain out-of-domain ability.} In most cases, models after rank-1 extraction exhibit varying degrees of significant performance degradation on non-mathematical reasoning tasks. This indicates that even a one-time rank-1 extraction can impair the model’s capabilities and knowledge. Therefore, although we find that periodic rank-1 substitution can mitigate potential implicit reward overfitting in RLVR, it cannot serve as an effective improvement strategy for RLVR. In other words, non-rank-1 components in RLVR play a crucial role in preserving the model’s non-reasoning abilities, thereby preventing catastrophic forgetting akin to that observed in supervised fine-tuning (SFT). 

\paragraph{Llama3.1-8B-Thinking-R1 seems to be a counterexample.} After rank-1 extraction, Llama3.1-8B-Thinking-R1-rank1 not only avoids performance degradation but also exhibits a significant performance improvement compared to Llama3.1-8B-Thinking-R1~\cite{jackrong2025llama318bthinkingr1}. This is because it is the only model in the table that has not been trained exclusively via RLVR. In fact, Llama3.1-8B-Thinking-R1 is derived from Llama-3.1-8B-Instruct through a three-stage training pipeline comprising cold-start SFT, GRPO, and CoT distillation SFT. Consequently, since SFT induces substantial shifts in the model’s parameter space, the extracted rank-1 component no longer encapsulates solely the reasoning capabilities attributable to pure RLVR, but also incorporates other knowledge. We posit that, under such circumstances, rank-1 extraction may help mitigate catastrophic forgetting. However, it remains challenging to guarantee that the model can preserve the reasoning abilities achieved through RL.

\input{table/effect-of-non-rank1}

\subsection{The Universal Spectral Pattern of $\Delta W$ in RLVR}

\begin{figure*}[ht]
    \centering
    \includegraphics[width=\linewidth]{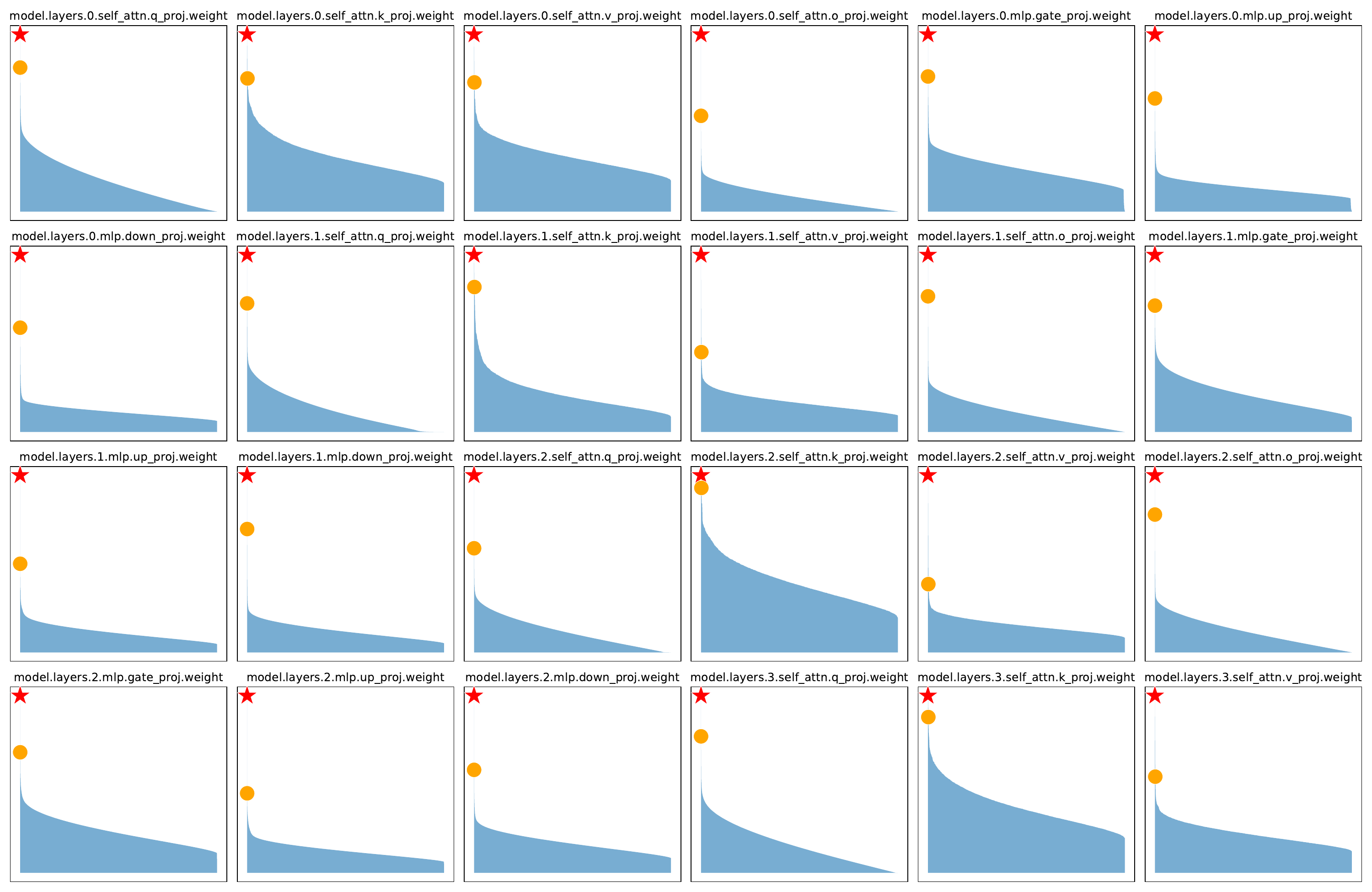}
    \caption{The distribution of the singular values of each linear layer update. The greatest singular value is marked by a red star, the second largest singular value is marked by a yellow circle. It's obvious that the heavy-tail part of non-rank-1 components take almost all energy of the parameter update, which echoes the need of the scaling factor in Section~\ref{rank1-preliminary}. Without scaling, the magnitude of the rank-1 component would be too small to come into effect. In another perspective, rank-1 component captures the desired direction within parameter space.}
    \label{fig:fig5-spectrum}
\end{figure*}

We performed SVD on the weight difference $\Delta W$ before and after RL training, sorting the singular values in descending order to obtain the singular spectrum shown in Figure \ref{fig:fig5-spectrum}. Through extensive verification, we identified a universal pattern in the singular spectrum of $\Delta W$: the first singular value $\sigma_1$ is notably large, followed by a rapid initial decay, while the subsequent singular values exhibit a uniform, quasi-linear rate of decay. This indicates that the tail singular vectors are systematically induced by the training process rather than being artifacts of random stochastic noise.

Furthermore, we observe that although the improvement in reasoning capability is primarily concentrated in the Rank-1 component, the effective rank of $\Delta W$ remains substantial. This implies that structurally, or numerically, the rank-1 component does not dominate the composition of the matrix relative to the non-rank-1 parts(in other words, rank-1 component doesn't bear the most energy of the weight). Synthesizing these findings, we propose a novel perspective: RL does not merely learn a strictly low-rank update but rather induces a highly structured spectral shape. Specifically, a dominant Rank-1 direction acts as a switch to activate reasoning capabilities, while RL systematically generates a series of linearly decaying secondary singular values. These non-Rank-1 components, characterized by low amplitude but consistent structure across layers and models, are responsible for safety, robustness, and subtle feature alignment, jointly refining the model's overall competence. 

\subsection{A Theoretical Perspective}
An outstanding study at ICLR~\cite{ren2024learningDynamics} lends strong support to the fact that the gradient of all RL objective functions can be formulated as $A * \nabla_\theta \log \pi_\theta$ in essence. In RLVR context, the model ultimately produces a policy  $ \pi_\theta(o_{i,t}\mid q,o_{i,<t}) $ . For  on-policy RL methods (GRPO/DAPO/GSPO family), the core term generally contains (we leave out unnecessary clip, min, etc.)
$$
J(\theta)=\mathbb{E}_{(q,a)\sim \mathcal{D},\{o_i\}\sim \pi_{\theta}}\big[\hat{A}_{i,t} \pi_\theta(o_{i,t}\mid q,o_{i,<t})\big].
$$
The gradient of the weights for this layer (via chain rule) becomes
$$
\nabla_W J = \mathbb{E}_{(q,a)\sim \mathcal{D},\{o_i\}\sim \pi_{\theta}}\Big[\hat{A}_{i,t} \nabla_W \pi_\theta(o_{i,t}\mid q,o_{i,<t})\Big]
$$
Let the layer input be  $ \boldsymbol x\in\mathbb{R}^{d_{\text{in}}} $ , weights  $ W\in\mathbb{R}^{d_{\text{out}}\times d_{\text{in}}} $ , and the output hidden state be $\boldsymbol h = W\boldsymbol x$. Define the backpropagated vector to the layer output as
$$
\boldsymbol \delta \triangleq \frac{\partial \pi_\theta(o_{i,t}\mid q,o_{i,<t})}{\partial \boldsymbol h} \in \mathbb{R}^{d_{\text{out}}}.
$$
Given  $ \boldsymbol h=W\boldsymbol x $ , matrix differential gives $\mathrm{d}\boldsymbol{h}=\mathrm{d}W\boldsymbol{x}$.
Thus
\begin{align*}
    \mathrm{d}\pi_\theta(o_{i,t}\mid q,o_{i,<t}) &= \boldsymbol{\delta}^T\mathrm{d}\boldsymbol{h} = \boldsymbol{\delta}^T(\mathrm{d}W\boldsymbol{x}) = \mathrm{tr}(\boldsymbol{\delta}^T\mathrm{d}W\boldsymbol{x}) = \mathrm{tr}(\boldsymbol{x}\boldsymbol{\delta}^T\mathrm{d}W) \\
    \implies \nabla_W\pi_\theta(o_{i,t}\mid q,o_{i,<t}) &= \boldsymbol{\delta}\boldsymbol{x}^T
\end{align*}
Substituting into $J$:
$$
\nabla_W J = \mathbb{E}_{(q,a)\sim \mathcal{D},\{o_i\}\sim \pi_{\theta}}\big[\hat{A}_{i,t} \boldsymbol \delta \boldsymbol x^T\big]
$$
For each token level probability, the gradient backpropagated to the weight matrix is a rank-1 matrix. The overall gradient is the average of these rank one matrices, which supports the rank-1 dominance in RLVR. 

\section{LoRA Unveils the Training Dynamics of RLVR}

\subsection{Experimental Perspective: Alignment in Output Space}
\begin{figure*}[t]
        \centering
    \includegraphics[width=\linewidth]{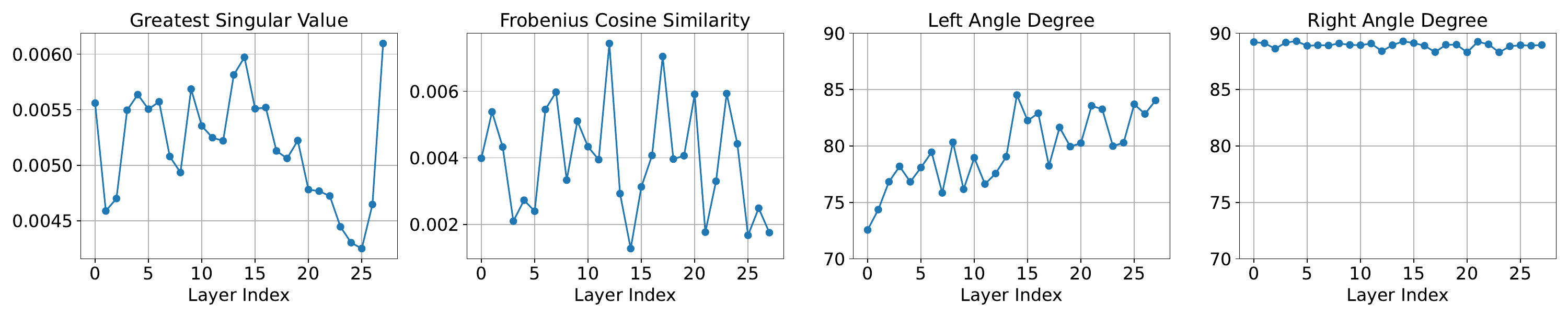}
    \caption{Layer-wise alignment analysis between $\Delta W_{\text{LoRA}}$ and $\Delta W^{(1)}$. From left to right: greatest singular value, Frobenius cosine similarity, left principal angle, and right principal angle across layers. The Frobenius cosine similarity remains near zero, indicating negligible global correlation in parameter space. In contrast, the subspace angle distributions show consistently smaller left principal angles than right ones, suggesting that the LoRA adapter, particularly its output-side component, tends to align more strongly with the left singular subspace during training.}
    \label{fig:fig3-alignment}
\end{figure*}

It comes as a natural idea to use LoRA on RLVR because the weight update formulation in LoRA bears a striking formal resemblance to the Rank-1 structure:
\begin{gather*}
    \Delta W_{LoRA}=\alpha A_{m\times r}B_{r\times n} \qquad\Delta W^{(1)}=\sigma_1\boldsymbol{u_1}\boldsymbol{v_1}^T
\end{gather*}
By setting $r=1$, the LoRA adapter matrices degenerate into vectors, rendering the two formulations mathematically identical. Consequently, we trained LoRA adapters(with $r=1$) using GRPO on the same dataset in Figure~\ref{fig:fig1}. To deeply probe the degree of alignment between LoRA updates and the true Rank-1 subspace, we employ two intrinsic metrics to quantify the approximation:

\textbf{Metric 1}: Frobenius Cosine Similarity, which evaluates the approximation of the weight update matrix from a holistic perspective:
\begin{gather*}
    \mathrm{Alignment}=\dfrac{\Vert \Delta W_{LoRA} \odot \Delta W^{(1)}\Vert_F}{\Vert \Delta W_{LoRA}\Vert_F \cdot \Vert \Delta W^{(1)}\Vert_F}
\end{gather*}

\textbf{Metric 2}: Principal Angles (Left and Right), which measure the alignment between the constituent subspaces (components) of the update matrices.
\begin{gather*}
    \theta_{\mathrm{left}}=\arccos{{\left|\dfrac{A_{m\times 1}^T\boldsymbol{u_1}}{\Vert A_{m\times 1}\Vert_F}\right|}}\\
    \theta_{\mathrm{right}}=\arccos{{\left|\dfrac{B_{1\times n}\boldsymbol{v_1}}{\Vert B_{1\times n}\Vert_F}\right|}}
\end{gather*}
As illustrated in Figure \ref{fig:fig3-alignment}, we find that from a holistic view, the Frobenius cosine similarity between $\Delta W_{LoRA}$ and $\Delta W^{(1)}$ is negligible, suggesting a lack of correlation in the parameter space. However, the distribution of subspace angles reveals a noteworthy phenomenon: the left principal angle is consistently smaller than the right. This indicates that during training, the LoRA adapter (specifically the output-side component $A$) exhibits a stronger tendency to align with the left singular vectors compared to the input side.

We attribute this to the distinct semantic information encoded within the input and output spaces. Geometrically, for a given weight matrix $W$, the left singular vectors(orthonormal eigenvectors of $WW^T$) form an orthonormal basis for its column space, constituting a representation of the layer's output, which is essentially the model's expressivity or sampling policy. Since RLVR fundamentally optimizes the sampling strategy or output representation, the shared semantic coordinate system on the output side is more easy to alignment. Conversely, the right singular vectors reside in the input-side feature combination space, which is highly contingent upon upstream activations and the contextual distribution. RL updates on the input side involve complex feature reorganization rather than simple directional adjustments, resulting in a dispersed input space that is difficult to align via simple low-rank matrices. In short, the left and right singular vectors exhibit asymmetric alignability in RLVR.

\subsection{Theoretical Proof}
This can also be explained from a theoretical perspective on gradient analysis. Consider a linear layer
$$
\boldsymbol{y} = W \boldsymbol{x},\quad W\in\mathbb{R}^{d_{\text{out}}\times d_{\text{in}}}, \boldsymbol{x}\in\mathbb{R}^{d_{\text{in}}}
$$
When LoRA of rank-1 is applied, the weight of this linear layer is:
$$
W' = W + \Delta W,\qquad \Delta W = \alpha \boldsymbol{a} \boldsymbol{b}^T
$$
where $a\in\mathbb{R}^{d_{\text{out}}},\quad b\in\mathbb{R}^{d_{\text{in}}},\quad \alpha>0$. The output $\boldsymbol{y}$ of this layer is:
$$
\boldsymbol{y}' = (W+\alpha \boldsymbol{a} \boldsymbol{b}^T)x = Wx + \alpha \boldsymbol{a} (\boldsymbol{b}^T x).
$$
Suppose the loss function is $\mathcal{L}$, we define the backpropagation vector $\boldsymbol{g}$ to this layer:
$$
\boldsymbol g \triangleq \frac{\partial \mathcal{L}}{\partial \boldsymbol y'}\in\mathbb{R}^{d_{\text{out}}}.
$$
Deriving gradients for the two vectors of rank-1 LoRA. First, differentiate with respect to $\Delta W$:
$$\frac{\partial \mathcal{L}}{\partial \Delta W} = \frac{\partial \mathcal{L}}{\partial \boldsymbol y'}\frac{\partial \boldsymbol y'}{\partial \Delta W}
= \boldsymbol g \boldsymbol x^T$$
Then, using the chain rule for $\Delta W=\alpha \boldsymbol a \boldsymbol b^T$:
\begin{gather*}
    \frac{\partial \mathcal{L}}{\partial \boldsymbol a}
= \alpha \frac{\partial \mathcal{L}}{\partial \Delta W} \boldsymbol b
= \alpha (\boldsymbol g \boldsymbol x^T)\boldsymbol b
= \alpha \boldsymbol g (\boldsymbol x^T \boldsymbol b) \\
\frac{\partial \mathcal{L}}{\partial\boldsymbol  b}
= \alpha \left(\frac{\partial \mathcal{L}}{\partial \Delta W}\right)^T \boldsymbol a
= \alpha (\boldsymbol x \boldsymbol g^T)\boldsymbol a
= \alpha \boldsymbol x (\boldsymbol g^T \boldsymbol a)
\end{gather*}
These are the root cause of the geometric asymmetry in gradients. The direction of $\nabla_{\boldsymbol a}$ is primarily determined by the output-side backpropagated error $\boldsymbol g$, scaled only by a scalar $(\boldsymbol x^T \boldsymbol b)$. In other words, the information of $\boldsymbol x$ is compressed to a scalar here. The direction of $\nabla_{\boldsymbol b}$ is primarily determined by the input activation $\boldsymbol x$, scaled only by a scalar $(\boldsymbol g^T \boldsymbol a)$. Therefore, as long as the RL signal causes $g$ to exhibit strong alignable directions during training, $a$ will be easier to align, whereas $x$ is often highly context-dependent, widely distributed, and directionally scattered, making $b$ difficult to align , especially in low-rank settings.

%% file: table/effect-of-non-rank1.tex
\begin{table}[t]
    \centering
    \small 
    \setlength\tabcolsep{10pt} 
    \renewcommand{\arraystretch}{1.3} 

    \begin{tabular}{l | c | c | c | c}
        \toprule
        \toprule
        \textbf{\underline{Model}} & \multicolumn{4}{c}{\textbf{\underline{Benchmark}}} \\
        \cmidrule(lr){2-5}
         & \textbf{IF-Eval} & \textbf{MMLU} & \textbf{MMLU-Pro} & \textbf{SafetyBench} \\
        \midrule
        \rowcolor{basiccolor} \multicolumn{5}{c}{\textbf{Instruction-tuned or RLVR-tuned Model}} \\
        \midrule
        Qwen2.5-7B-Instruct &  $78.66\pm 1.2$  & $74.64\pm 1.8$  &  $45.01\pm 1.4$  & $78.34 \pm 1.9$  \\
        Qwen2.5-7B-R1 &  $78.18\pm 1.1$  &  $74.29 \pm 1.5$  &  $52.93\pm 2.1$  & $77.41 \pm 1.7$  \\
        DeepMath-Zero-7B &  \color{red}$38.73\pm 1.9$  &  $77.95\pm 1.2$  &  $60.07\pm 1.4$  & $82.32 \pm 1.1$  \\
        Qwen2.5-32B-DAPO & $53.36\pm 0.7$  &  $86.51\pm 0.8$  &  $68.32\pm 1.1$  &  $86.61 \pm 0.9$  \\
        Qwen3-8B-DAPO-Math &  $52.04\pm 1.4$  &  $76.55\pm 2.0$  &  $62.40\pm 1.6$& $54.87 \pm 1.3$  \\
        \rowcolor{headercolor} Llama3.1-8B-Thinking-R1 &  $60.91\pm 1.6$  &  $61.54\pm 1.7$  & $35.38\pm 1.5$ & $45.22 \pm 2.2$  \\
        \midrule
        \rowcolor{complexcolor} \multicolumn{5}{c}{\textbf{Rank-1 Model}} \\
        \midrule
        Qwen2.5-7B-R1-rank1 &  $74.64\pm 1.3$ & $71.15\pm 1.6$ & $49.57\pm 1.8$ & $73.22 \pm 1.9$    \\
        DeepMath-Zero-7B-rank1 & \color{red}$31.57\pm 1.8$ & $74.81\pm 1.4$ & $57.24\pm 1.6$ & $78.50 \pm 1.3$   \\
        Qwen2.5-32B-DAPO-rank1 & $46.73\pm 0.8$ & $83.43\pm 0.9$ & $64.50\pm 1.2$ & $82.16 \pm 1.1$   \\
        Qwen3-8B-DAPO-Math-rank1 &  $55.04\pm 1.5$  &  $73.94\pm 1.4$  &  $58.89\pm 1.3$  & $51.08 \pm 1.8$  \\
        \rowcolor{headercolor} Llama3.1-8B-Thinking-R1-rank1 &  $57.67\pm 1.1$  &  $67.79\pm 1.9$  &  $41.88\pm 1.2$  & $66.14 \pm 2.0$  \\
        \bottomrule
        \bottomrule
    \end{tabular}
    \vspace{1mm}
    \caption{Out-of-domain ability of models before/after rank-1 extraction, including instruction-following(IF-Eval), world knowledge(MMLU, MMLU-Pro) and safety(SafetyBench). Note that DeepMath-Zero-7B~\cite{he2025deepmath} performs especially bad on IF-Eval as the post-training of it utilized Open-Reasoner~\cite{hu2025openReasoner} chat template instead of default chat template, according to its technical report.}
    \label{tab:tab1-effect-of-non-rank1}
\end{table}

%% file: 3-experiment.tex
\section{Experiments}

\paragraph{Experiment of Figure \ref{fig:fig2-overfit}.} We trained Qwen2.5-7B-Instruct on Countdown-3to4~\cite{tinyzero} dataset using GRPO method. At each training step, we sample 8 model responses for each question and there are 32 questions in a rollout batch. During rollout, top\_p for sampling is 0.95. Learning rate is 5e-6. We use AdamW~\cite{adam2014method,loshchilov2017adamw} optimizer with $\beta_1=0.9$ and $\beta_2=0.999$. It takes 6 hours to train on a single RTX Pro 6000 GPU. We sampled the last 1000 question out as the test set. For experiment in mid-right figure, we trained Llama3.1-8B-Instruct~\cite{grattafiori2024llama3herdmodels} on GSM8K~\cite{cobbe2021gsm8k} dataset. At each training step, we sample 8 model responses also for each question. However, gradient accumulation is employed here. There are only 2 questions in a rollout batch thus we perform a gradient update every 16 rollout steps. top\_p is also 0.95. A gradient clipping of 1.0 is applied. It takes approximately 20 hours on a single RTX Pro 6000 GPU. Note that both two experiment nearly consumed all VRAM, so at least 96GB VRAM is needed to reproduce the result.

\paragraph{Experiment of Figure \ref{fig:fig4-safety}.} We sampled the first 10k questions in SafetyBench-test-en.json~\cite{zhang2024safetybench}, and evaluate models on a single NVIDIA H100 GPU using VLLM engine~\cite{kwon2023PagedAttention}. Hyperparameters are top\_p=0.95, temperature=0.9, batch\_size=256, max\_tokens=10000.

\paragraph{Experiment of Table~\ref{tab:tab1-effect-of-non-rank1}.} For IF-Eval~\cite{zhou2023instructionfollowingevaluationlargelanguage}, the accuracy in the table is the instruction-level accuracy under strict mode. We set temperature=top\_p=0.95, and max\_tokens=10000. For MMLU~\cite{hendryckstest2021MMLU}, We set temperature=top\_p=0.95, and max\_tokens=10000. For MMLU-Pro~\cite{wang2024mmluPro}, we set temperature=0.9, top\_p=0.95, max\_tokens=5000. For SafetyBench~\cite{zhang2024safetybench}, we sampled the first 10k questions and we set temperature=0.9, top\_p=0.95. Evaluation takes a single NVIDIA H100 GPU using VLLM engine~\cite{kwon2023PagedAttention}.

\paragraph{Experiment of Figure~\ref{fig:fig3-alignment}}. We follow the configuration of \cite{schulman2025lorawithoutregret}, training Qwen2.5-7B-Instruct~\cite{qwen2025qwen25technicalreport} on a subset of OpenThought-1.2M~\cite{guha2025openthoughts} using Verl~\cite{sheng2025hybridflow} framework. The training takes 4 NVIDIA H100 80GB GPUs.

%% file: 4-related-work.tex
\section{Related Works}
\paragraph{Reinforcement Learning for LLM.} Before the emergence of reasoning-capable models such as OpenAI's o1, reinforcement learning was primarily employed in RLHF to improve instruction-following and alignment with human preferences. More recently, RL with Verifiable Rewards has been proposed as an effective strategy to enhance reasoning in domains such as mathematics and programming. OpenAI's ol was the first to demonstrate that RL can incentivize large-scale reasoning, inspiring subsequent models such as DeepSeek-R1\cite{guo2025deepseek-r1}, and Qwen3~\cite{yang2025qwen3technicalreport} . Building on these advances, later approaches such as Dr.GRPO~\cite{liu2025drgrpo}, CISPO~\cite{minimax2025minimaxm1}, GFPO~\cite{shrivastava2025gfpo}, GMPO~\cite{zhao2025gmpo}, etc. have further broadened the landscape of RL-based reasoning.

\paragraph{Interpreting Reinforcement learning.} A recent study~\cite{cui2025entropymechanismreinforcementlearning} identified the
phenomenon of \textbf{entropy collapse} in reinforcement learning, where rapid early convergence causes the model to become overly confident, prematurely degrading its exploratory capacity. A related study~\cite{shenfeld2025rlsrazoronlinereinforcement} observed in chain-of-thought reasoning that high-entropy tokens often act as branching points defining multiple potential reasoning paths.

%% file: 5-conclusion.tex
\section{Conclusion and Limitations}
This study observes, through Periodic Rank-1 Substitution, that while RLVR enhances training rewards, it does not necessarily yield corresponding improvements in test-set generalization, suggesting the presence of implicit reward overfitting. Furthermore, we analyze RLVR from the perspective of parameter spectral structure and find that its reasoning gains are primarily concentrated along the Rank-1 update direction. In contrast, the Rank-1 component predominantly serves to preserve general capabilities, such as safety, knowledge retention, and instruction-following. We further substantiate this finding using LoRA, demonstrating that vectors on the output side are more easy to alignment.

The limitations of our work are: due to limited computing resources, we have no opportunity to observe the behavior of models with more parameters. Furthermore, we don't explore whether our findings hold ture for new model architectures. These questions require further study to solve.

%% file: 6-appendix.tex
\appendix

\section{Why RLVR Algorithms Are the Same in Essence}
Let
$$
\hat A_i=\frac{R_i-\mu_R}{\sigma_R},\qquad \mu_R=\frac1G\sum_{j=1}^G R_j.
$$
GRPO objective:
$$
J_{\rm GRPO}(\theta)=\mathbb E\Bigg[
\frac1G\sum_{i=1}^G \frac1{|y_i|}\sum_{t=1}^{|y_i|}
\min\!\Big(r_{i,t}(\theta)\hat A_i,\ \mathrm{clip}(r_{i,t}(\theta),1-\epsilon,1+\epsilon)\hat A_i\Big)
-\beta D_{\rm KL}(\pi_\theta\|\pi_{\rm ref})
\Bigg],
$$

$$
r_{i,t}(\theta)=\frac{\pi_\theta(y_{i,t}\mid x,y_{i,<t})}{\pi_{\rm old}(y_{i,t}\mid x,y_{i,<t})}.
$$

DAPO:
\[
J_{\rm DAPO}(\theta)=\mathbb E\Bigg[
\frac1{\sum_j |y_j|}\sum_{i=1}^G\sum_{t=1}^{|y_i|}
\min\!\Big(r_{i,t}(\theta)\hat A_i,\ \mathrm{clip}(r_{i,t}(\theta),1-\epsilon_{\rm low},1+\epsilon_{\rm high})\hat A_i\Big)
\Bigg],
\]

Here, dynamic sampling per se is not a loss, it's a sampling strategy.
\[
0<\big|\{\,y_i\mid \mathrm{is\_equivalent}(a,y_i)\,\}\big|<G.
\]
overlong shaping:
\[
R_i=R_{\rm task}(y_i,a)+R_{\rm len}(y_i),
\]

\[
R_{\rm len}(y)=
\begin{cases}
0, & |y|\le L_{\max}-L_{\rm cache}\\
\dfrac{L_{\max}-L_{\rm cache}-|y|}{L_{\rm cache}}, & L_{\max}-L_{\rm cache}<|y|\le L_{\max}\\
-1, & |y|>L_{\max}.
\end{cases}
\]

GSPO objective:
\[
s_i(\theta)=\left(\frac{\pi_\theta(y_i\mid x)}{\pi_{\rm old}(y_i\mid x)}\right)^{1/|y_i|}
=\exp\!\left(\frac1{|y_i|}\sum_{t=1}^{|y_i|}\log\frac{\pi_\theta(y_{i,t}\mid x,y_{i,<t})}{\pi_{\rm old}(y_{i,t}\mid x,y_{i,<t})}\right),
\]
\[
J_{\rm GSPO}(\theta)=\mathbb E\Bigg[
\frac1G\sum_{i=1}^G
\min\!\Big(s_i(\theta)\hat A_i,\ \mathrm{clip}(s_i(\theta),1-\epsilon,1+\epsilon)\hat A_i\Big)
\Bigg].
\]

Regardless of the clip,and treat \(\pi_{\rm old}\) and  \(\hat A_i\) as stop-grad constants, then
\[
\nabla_\theta r_{i,t}(\theta)=r_{i,t}(\theta)\nabla_\theta\log \pi_\theta(y_{i,t}\mid x,y_{i,<t}),
\]
\[
\nabla_\theta s_i(\theta)=s_i(\theta)\frac1{|y_i|}\sum_{t=1}^{|y_i|}\nabla_\theta\log \pi_\theta(y_{i,t}\mid x,y_{i,<t}).
\]
all gradients update can be formulated as:
\[
\nabla_\theta J \approx \mathbb E\!\left[\sum_{i,t}\omega_{i,t}\,\nabla_\theta\log \pi_\theta(y_{i,t}\mid x,y_{i,<t})\right],
\]
In other words, this approach performs signed weighted Supervised Fine-Tuning (SFT) on the sampled tokens or responses. The primary distinction lies in the definition of the weight \(\omega\):
\begin{itemize}
    \item GRPO defines the weight as the token-level probability ratio multiplied by the advantage function.
    \item DAPO also operates at the token level but introduces an asymmetric clipping mechanism. Furthermore, it shifts the normalization for long samples from sample-average to token-mean, while reformulating \(R_i\) through dynamic sampling and length-based rewards.
    \item GSPO, in contrast, elevates the probability ratio to the sequence level. Consequently, all tokens within a single response share the same weight \(s_i\), which explains why GSPO more closely resembles whole-sequence weighted SFT.
\end{itemize}

\section{More Examples of Singular Spectrum}

\begin{figure}[h]
    \centering
    \includegraphics[width=\linewidth]{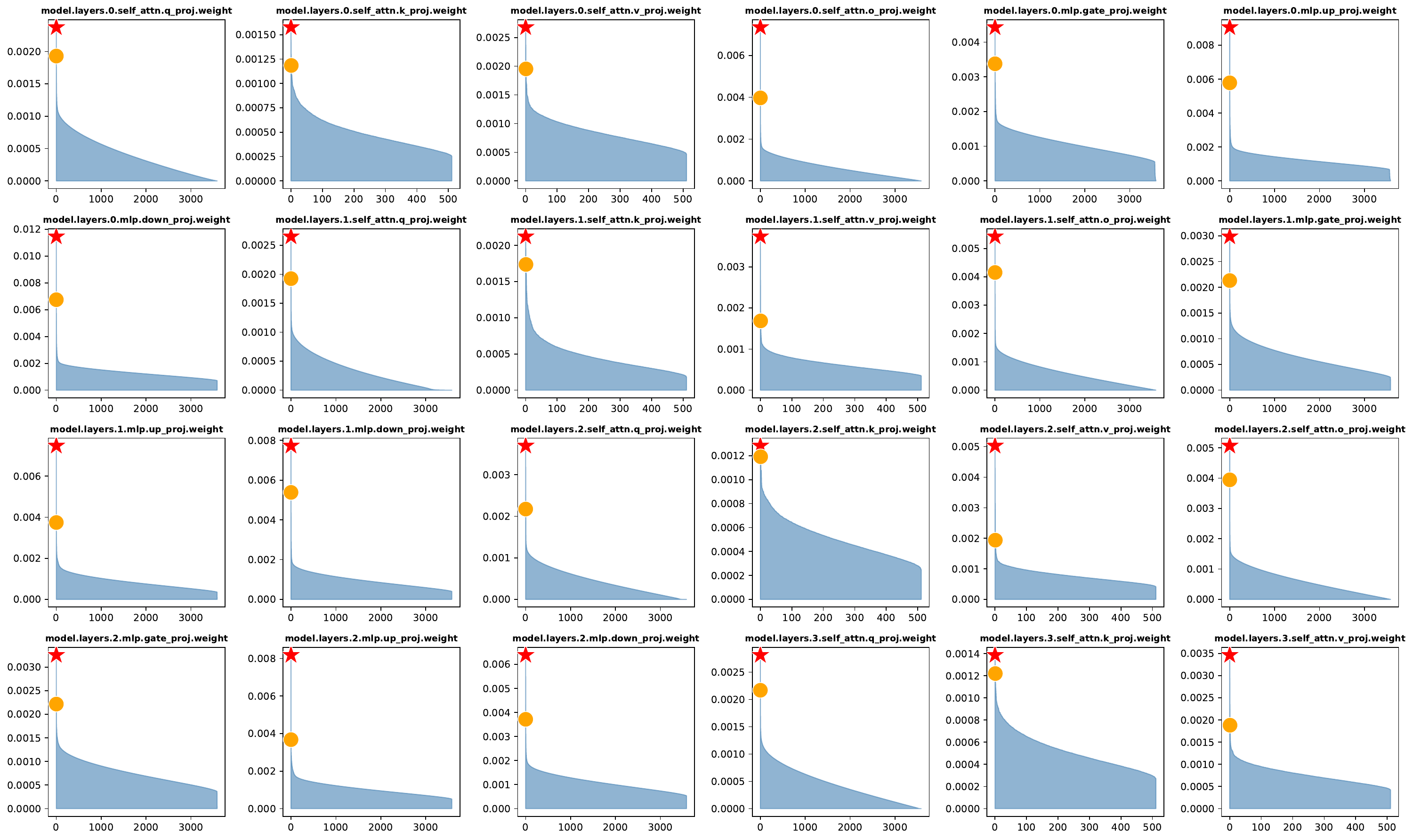}
    \caption{step10}
    \label{fig:placeholder}
\end{figure}
\begin{figure}
    \centering
    \includegraphics[width=\linewidth]{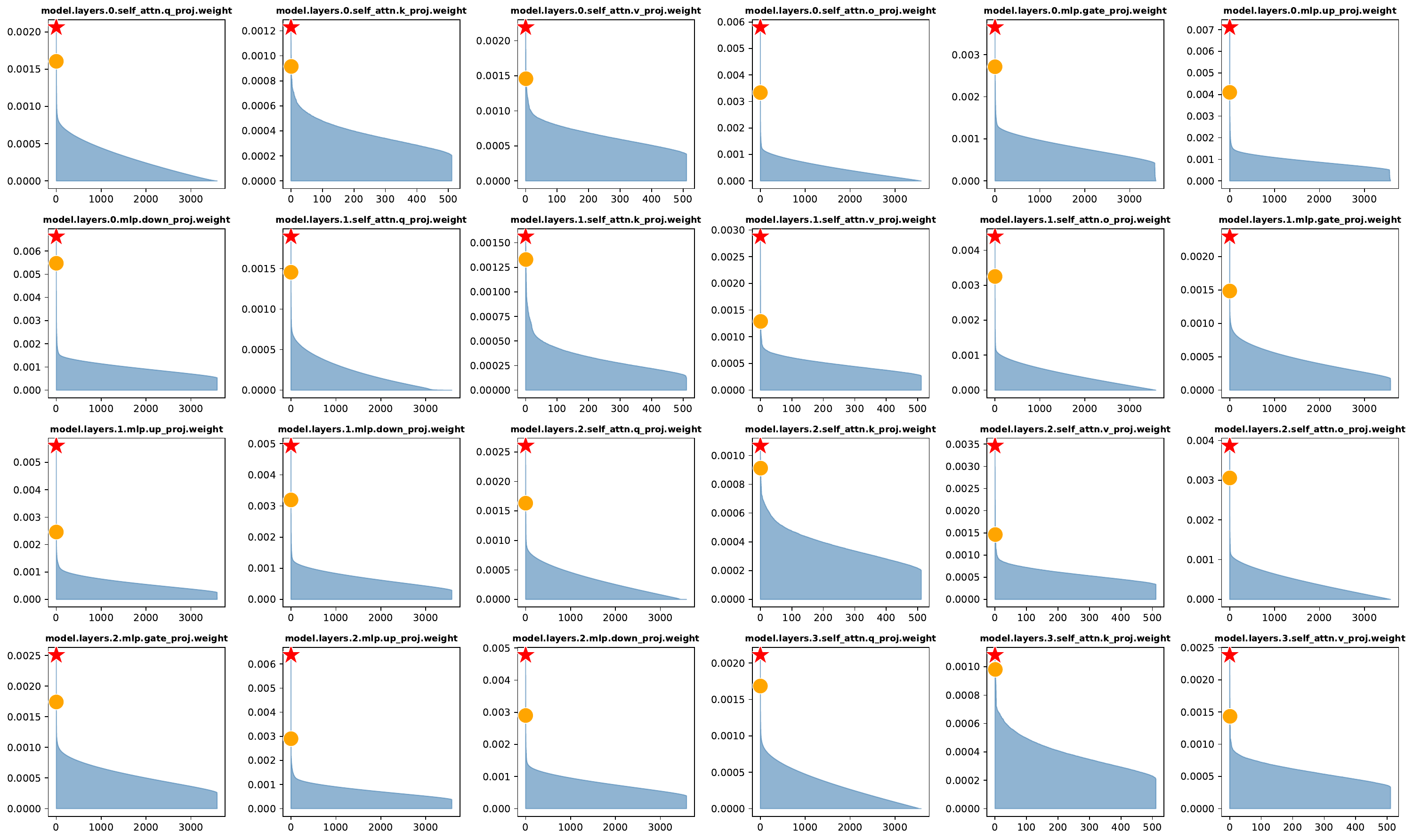}
    \caption{step20}
    \label{fig:placeholder}
\end{figure}

\begin{figure}
    \centering
    \includegraphics[width=\linewidth]{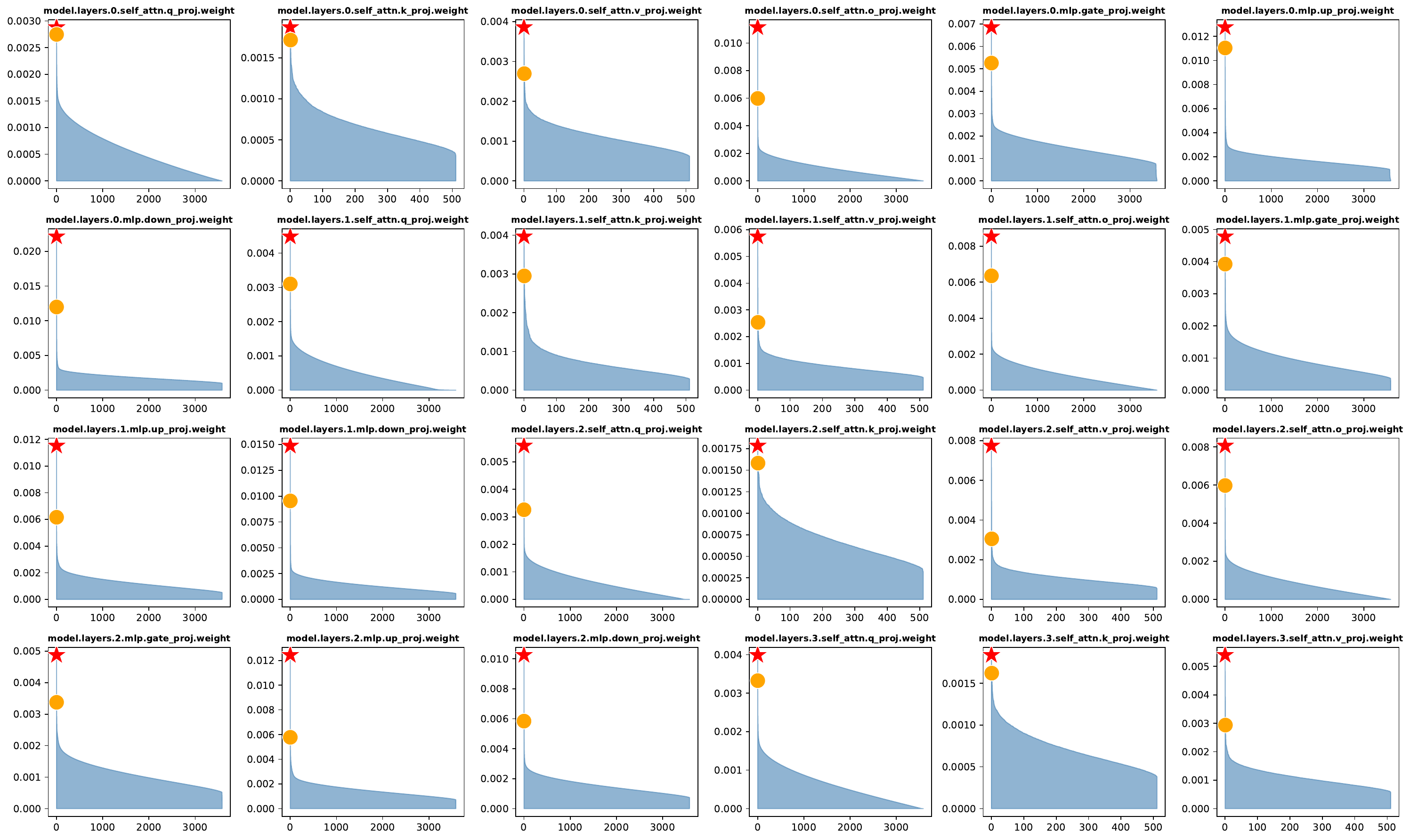}
    \caption{step30}
    \label{fig:placeholder}
\end{figure}

\begin{figure}
    \centering
    \includegraphics[width=\linewidth]{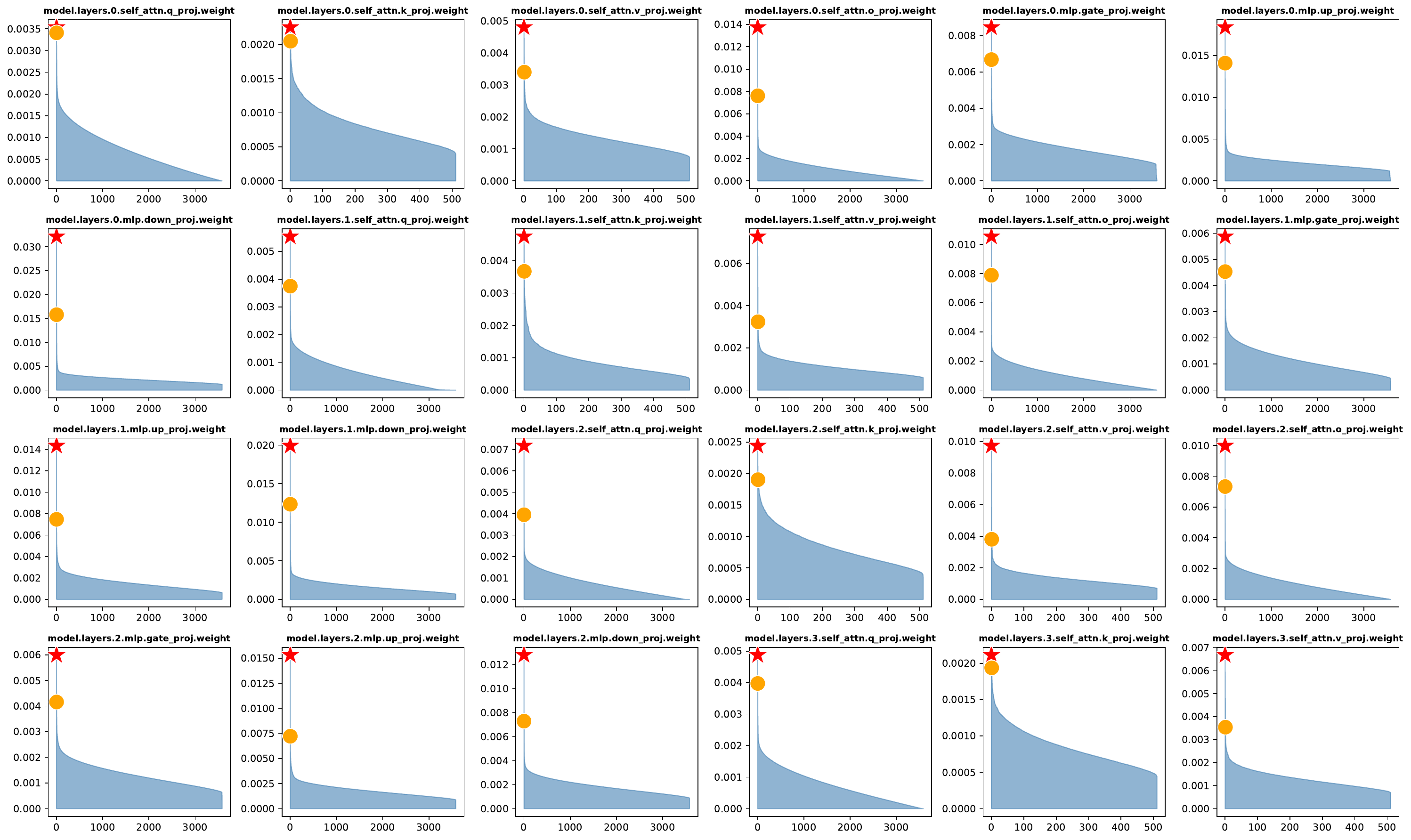}
    \caption{step40}
    \label{fig:placeholder}
\end{figure}

\begin{figure}
    \centering
    \includegraphics[width=\linewidth]{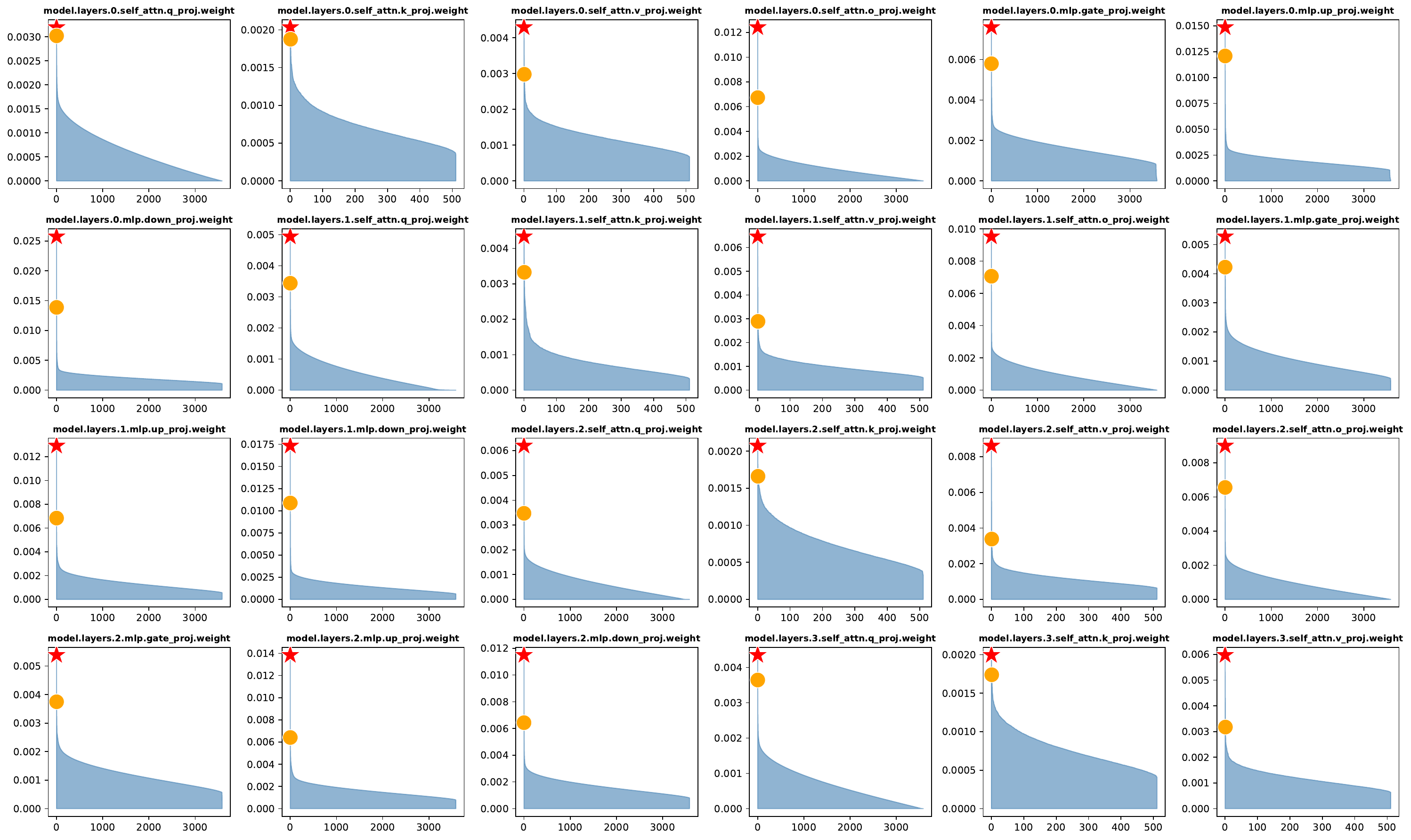}
    \caption{step50}
    \label{fig:placeholder}
\end{figure}

\begin{figure}
    \centering
    \includegraphics[width=\linewidth]{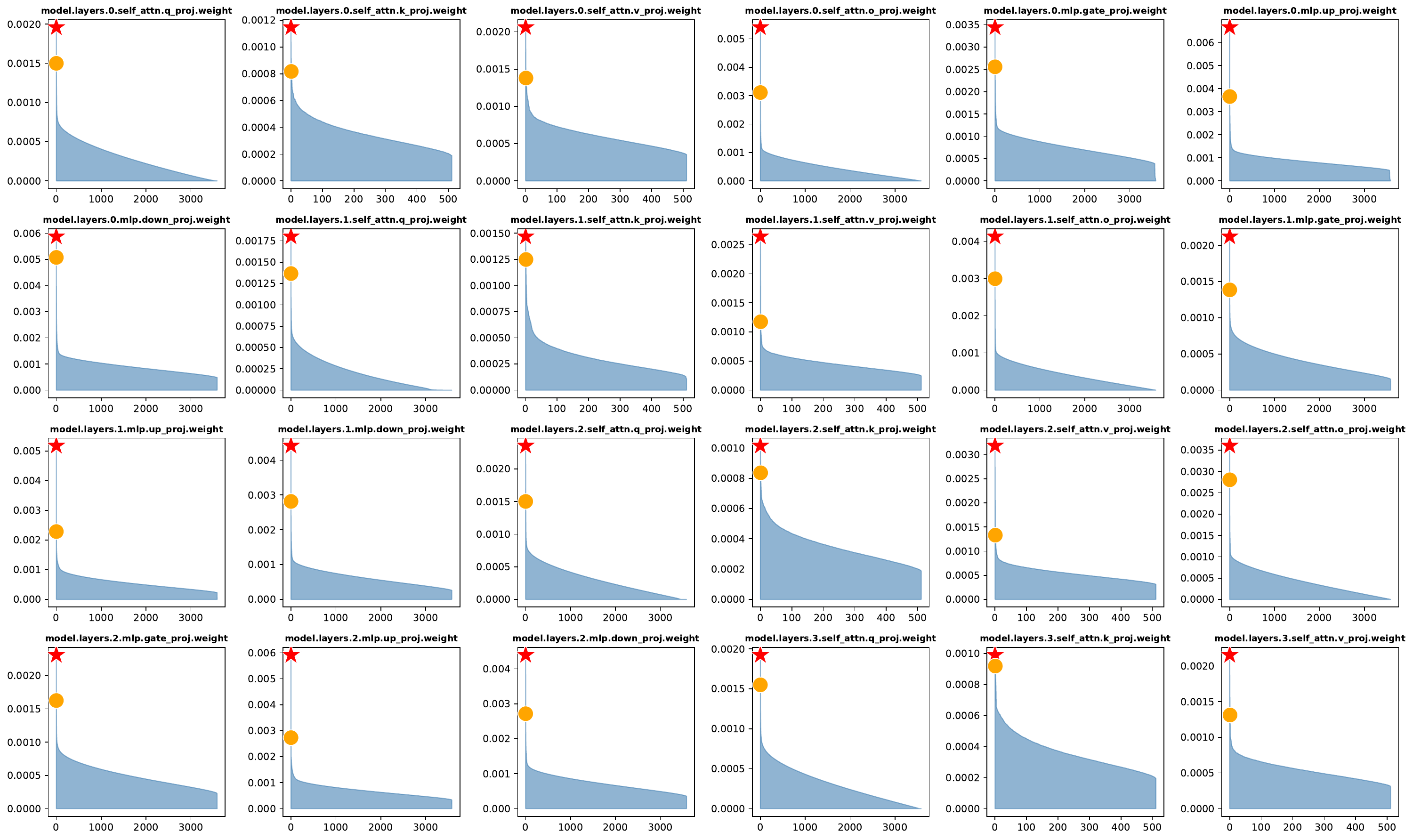}
    \caption{step60}
    \label{fig:placeholder}
\end{figure}

\begin{figure}
    \centering
    \includegraphics[width=\linewidth]{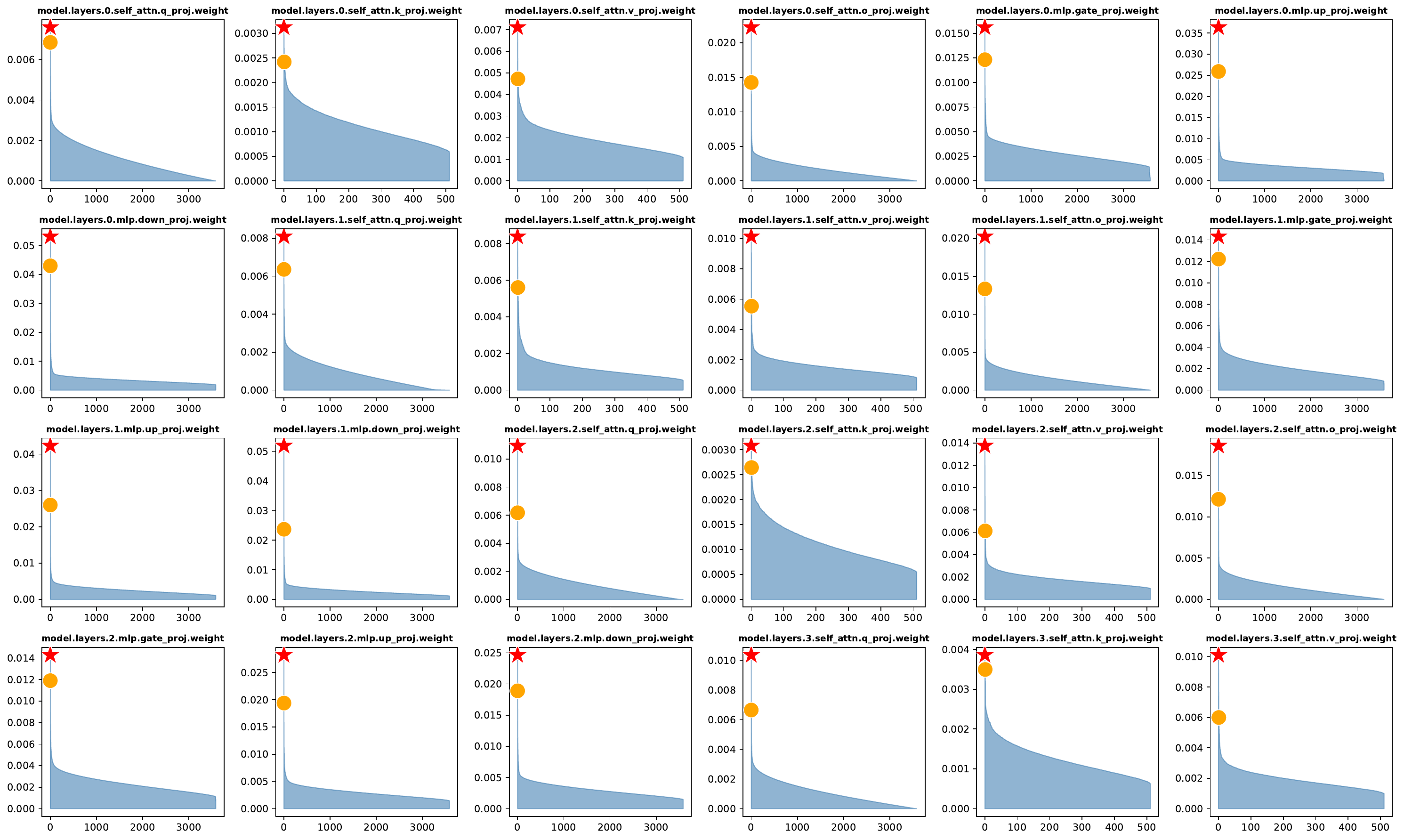}
    \caption{step70}
    \label{fig:placeholder}
\end{figure}

\begin{figure}
    \centering
    \includegraphics[width=\linewidth]{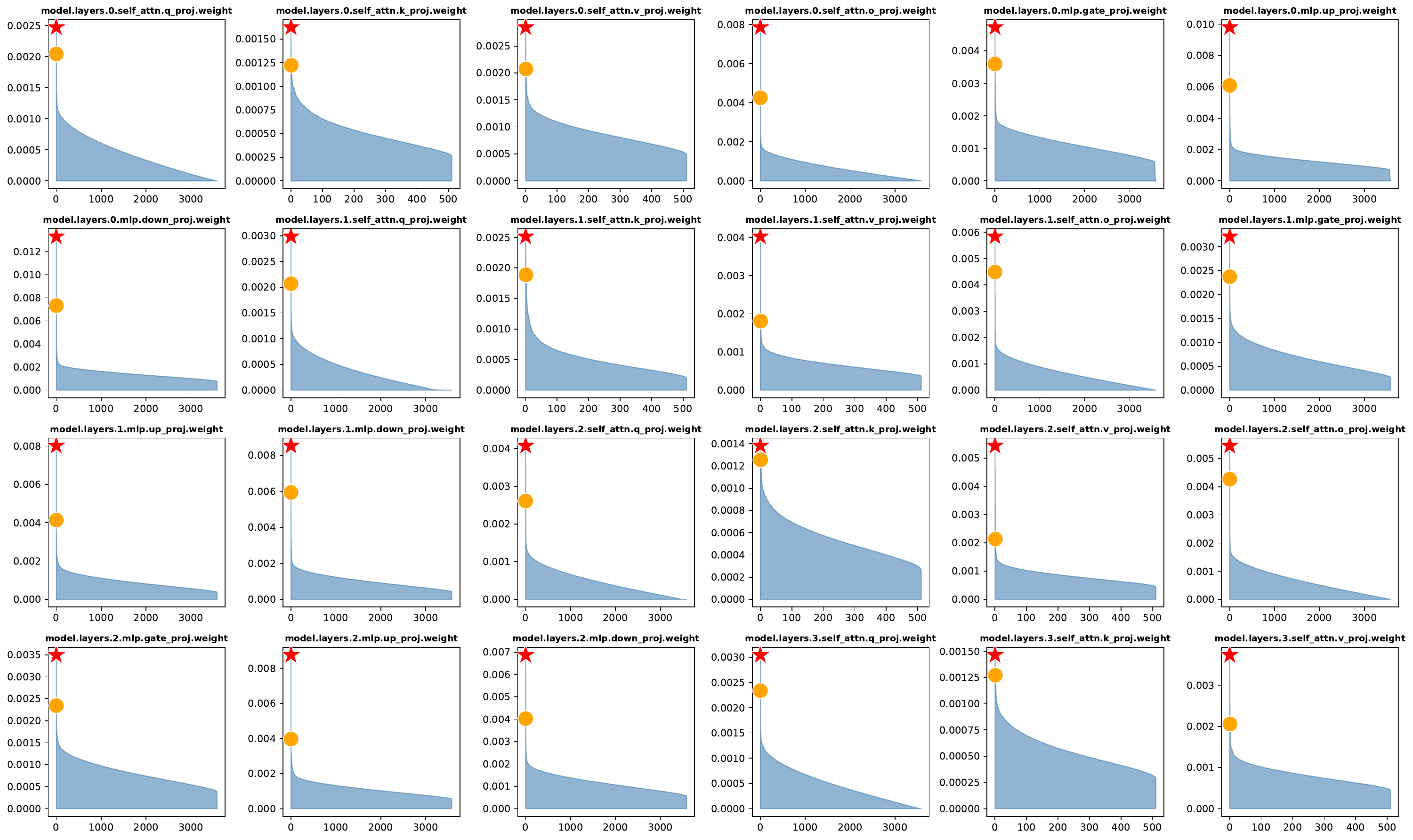}
    \caption{step80}
    \label{fig:placeholder}
\end{figure}

\clearpage
\section{More Results for Singular Spectra}
As shown in Table~\ref{tab:svd-dominant-linear-layers}, deep layers(near the output side) tend to have clearer pattern of "leading spike followed by a heavy tail".
\begin{table}[h]
\centering
\small
\begin{tabular}{lccccccc}
\toprule
Layer & $\sigma_1$ & $\sigma_2$ & $\sigma_3$ & $\sigma_4$ & $\sigma_5$ & $\sigma_6$ & $\sigma_7$ \\
\midrule
\texttt{layers.0.mlp.gate\_proj} & 100.00\% & 42.40\% & 41.85\% & 41.58\% & 41.43\% & 40.63\% & 39.84\% \\
\texttt{layers.0.mlp.up\_proj} & 100.00\% & 63.97\% & 63.85\% & 63.75\% & 63.66\% & 63.61\% & 63.58\% \\
\texttt{layers.1.mlp.up\_proj} & 100.00\% & 44.48\% & 40.70\% & 39.56\% & 38.76\% & 36.81\% & 36.62\% \\
\texttt{layers.2.mlp.up\_proj} & 100.00\% & 45.38\% & 44.93\% & 44.85\% & 44.77\% & 44.68\% & 44.62\% \\
\texttt{layers.6.mlp.up\_proj} & 100.00\% & 64.69\% & 64.64\% & 64.58\% & 64.53\% & 64.45\% & 64.34\% \\
\texttt{layers.19.mlp.up\_proj} & 100.00\% & 67.90\% & 67.24\% & 65.72\% & 65.41\% & 65.37\% & 65.32\% \\
\texttt{layers.21.mlp.up\_proj} & 100.00\% & 69.08\% & 67.27\% & 67.25\% & 67.16\% & 67.09\% & 67.05\% \\
\texttt{layers.22.mlp.down\_proj} & 100.00\% & 64.60\% & 57.21\% & 56.70\% & 56.48\% & 55.97\% & 55.74\% \\
\texttt{layers.22.mlp.up\_proj} & 100.00\% & 68.26\% & 67.39\% & 67.13\% & 67.04\% & 67.02\% & 66.91\% \\
\texttt{layers.22.self\_attn.o\_proj} & 100.00\% & 68.23\% & 66.25\% & 66.02\% & 65.63\% & 64.81\% & 64.46\% \\
\texttt{layers.23.mlp.down\_proj} & 100.00\% & 63.83\% & 58.42\% & 58.27\% & 58.20\% & 58.15\% & 58.08\% \\
\texttt{layers.23.self\_attn.o\_proj} & 100.00\% & 49.44\% & 48.39\% & 47.25\% & 47.18\% & 46.55\% & 46.32\% \\
\texttt{layers.24.mlp.down\_proj} & 100.00\% & 61.17\% & 59.56\% & 59.51\% & 59.45\% & 59.38\% & 59.34\% \\
\texttt{layers.24.self\_attn.o\_proj} & 100.00\% & 42.64\% & 39.82\% & 39.71\% & 39.71\% & 39.62\% & 39.53\% \\
\texttt{layers.25.mlp.down\_proj} & 100.00\% & 68.28\% & 66.29\% & 66.07\% & 65.76\% & 65.54\% & 65.50\% \\
\texttt{layers.25.self\_attn.o\_proj} & 100.00\% & 54.14\% & 48.65\% & 46.65\% & 46.48\% & 46.43\% & 46.34\% \\
\texttt{layers.26.self\_attn.o\_proj} & 100.00\% & 43.54\% & 41.27\% & 40.71\% & 40.64\% & 40.53\% & 40.43\% \\
\texttt{layers.27.mlp.down\_proj} & 100.00\% & 48.49\% & 44.82\% & 43.14\% & 39.28\% & 38.82\% & 37.78\% \\
\texttt{layers.27.mlp.gate\_proj} & 100.00\% & 65.37\% & 61.14\% & 60.30\% & 57.03\% & 56.24\% & 55.29\% \\
\texttt{layers.27.mlp.up\_proj} & 100.00\% & 62.04\% & 61.90\% & 61.67\% & 61.09\% & 60.56\% & 60.43\% \\
\bottomrule
\end{tabular}
\caption{Linear layers of Qwen2.5-7B-R1 with $\sigma_2 / \sigma_1 \le 0.70$; entries are percentages normalized by each layer's largest singular value.}
\label{tab:svd-dominant-linear-layers}
\end{table}